\pdfoutput=1

\documentclass[11pt]{article}

\usepackage[final]{acl}

\usepackage{times}
\usepackage{latexsym}

\usepackage[T1]{fontenc}

\usepackage[utf8]{inputenc}

\usepackage{microtype}

\usepackage{inconsolata}

\usepackage{graphicx}
\usepackage{mathtools}
\usepackage{paralist}
\usepackage{subcaption}
\usepackage{multirow}
\usepackage{booktabs}
\usepackage{tabularx}
\usepackage{cleveref}
\usepackage{caption}
\usepackage{adjustbox}
\usepackage{xspace}
\usepackage{arydshln}
\usepackage{makecell}
\usepackage{enumitem}
\usepackage{xurl}
\setlist[enumerate]{itemsep=0mm}

\graphicspath{ {./images/} }

\author{Min-Hsuan Yeh\textsuperscript{1}~~~~Ruyuan Wan\textsuperscript{2}~~~~Ting-Hao `Kenneth' Huang\textsuperscript{2}\\
\textsuperscript{1}University of Wisconsin-Madison, Madison, WI, USA. \\
  \texttt{samuelyeh@cs.wisc.edu}\\
  \textsuperscript{2}The Pennsylvania State University, University Park, PA, USA. \\
  \texttt{\{rjw6289,txh710\}@psu.edu}
}

\usepackage{xspace}
\usepackage{comment}
\usepackage{xcolor}
\usepackage{multirow}
\usepackage{booktabs}
\usepackage{multirow}
\usepackage{cleveref}

\usepackage{soul}

\title{\dataset: A Dataset of News \underline{Co}mments with \underline{Co}mmon \underline{Lo}gical \underline{Fa}llacies Written by LLM-Assisted Crowds}

\newcommand{\kenneth}[1]{{\small\color{blue}{\bf\xspace[#1 -Kenneth]}}}
\newcommand{\samuel}[1]{{\small\color{magenta}{\bf\xspace[#1 -Samuel]}}}

\newcommand{\eg}{{\it e.g.}\xspace}
\newcommand{\ie}{{\it i.e.}\xspace}

\newcommand{\dataset}{\mbox{\textsc{CoCoLoFa}}\xspace}

\newcommand{\ncomment}{7,706\xspace}
\newcommand{\nnews}{648\xspace}
\newcommand{\nworker}{143\xspace}

\newcommand{\QOne}{Q3\xspace}
\newcommand{\QTwo}{Q4\xspace}
\newcommand{\QThree}{Q2\xspace}
\newcommand{\QFour}{Q1\xspace}

\begin{document}
\maketitle

\begin{abstract}



Detecting logical fallacies in texts can help users spot argument flaws, but automating this detection is not easy.
Manually annotating fallacies in large-scale, real-world text data to create datasets for developing and validating detection models is costly.
This paper introduces \dataset, the largest known English logical fallacy dataset, containing \ncomment comments for \nnews news articles, with each comment labeled for fallacy presence and type.
We recruited \nworker crowd workers to write comments embodying specific fallacy types (\eg, slippery slope) in response to news articles.
Recognizing the complexity of this writing task, we built an LLM-powered assistant into the workers' interface to aid in drafting and refining their comments.
Experts rated the writing quality and labeling validity of \dataset as high and reliable.
BERT-based models fine-tuned using \dataset achieved the highest fallacy detection (F1=0.86) and classification (F1=0.87) performance on its test set, outperforming the state-of-the-art LLMs.
Our work shows that combining crowdsourcing and LLMs enables us to more effectively construct datasets for complex linguistic phenomena that crowd workers find challenging to produce on their own. 
\dataset is public at \href{https://cocolofa.org/}{CoCoLoFa.org/}.

\end{abstract}



\section{Introduction\label{sec:introduction}}

\begin{figure}[t]
    \centering
    \includegraphics[width=\columnwidth]{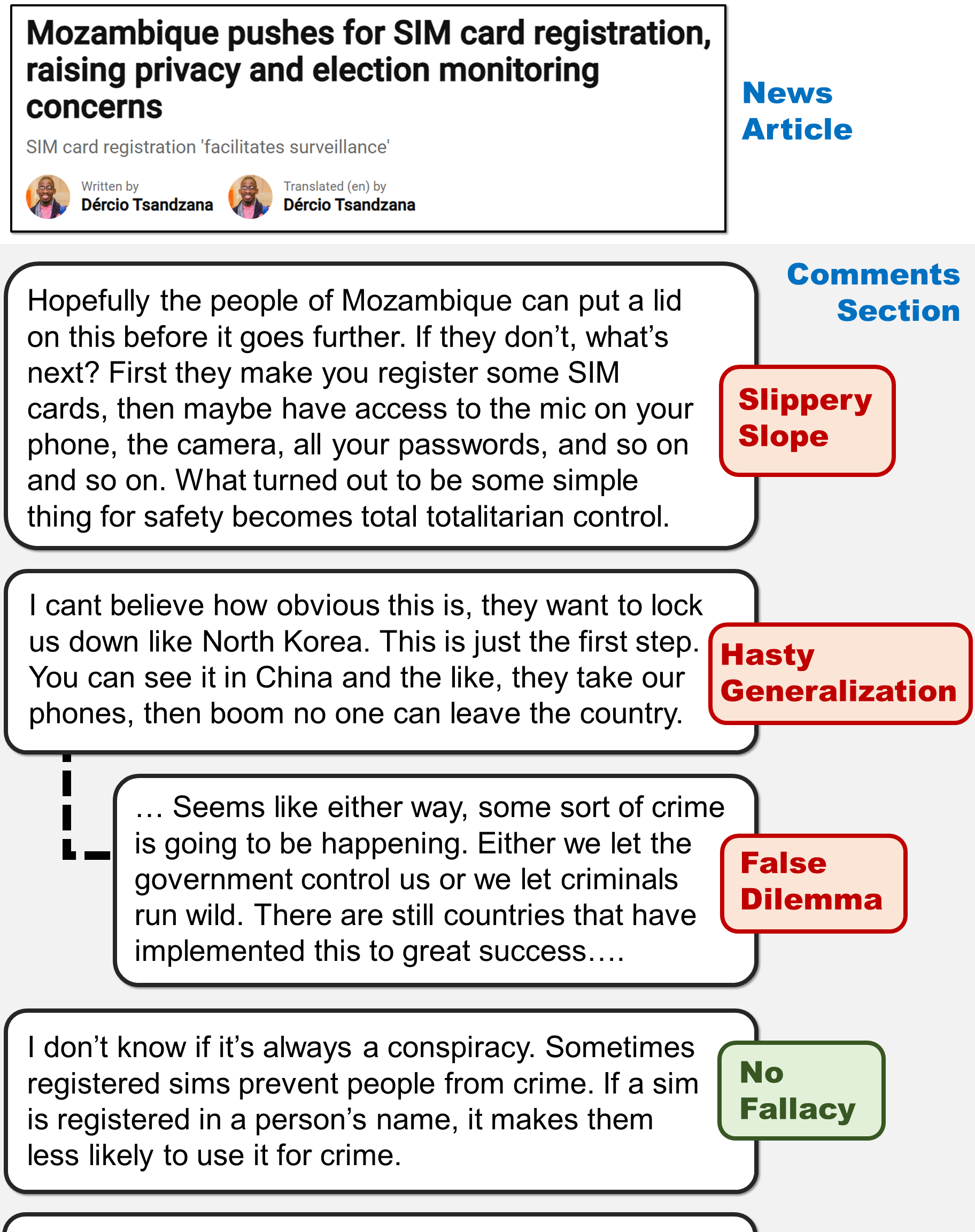}
    \caption{Examples from \dataset. For each news article, we hired crowd workers to form a thread of comment. Each worker was assigned to write a comment with a specific type of logical fallacy (or a neutral argument) in response to the article.
    }
    \label{fig:examples}
\end{figure}

\begin{table*}[t]
    \centering
    \small
    \begin{tabular}{llccccccc}
        \toprule
        Dataset & Genre & \# Topics & \# Fallacies & \makecell[b]{Total\\\# Item} & \makecell[b]{\# Neg.\\Item}  & \makecell[b]{\# Sent. \\per Item} & \makecell[b]{\# Tokens\\ per Item} & \makecell[b]{Vocab.\\Size}\\
        \midrule
        \makecell[cl]{\textsc{Logic}\\ \citep{Jin_Lalwani_Vaidhya_Shen_Ding_Lyu_Sachan_Mihalcea_Schoelkopf_2022}} & \makecell[cl]{Quiz\\ questions} & N/A & 13 & 2,449 & 0 & 1.92 & 31.20 & 7,624\\
        \\[-0.8em]
        \makecell[cl]{\textsc{LogicClimate}\\ \citep{Jin_Lalwani_Vaidhya_Shen_Ding_Lyu_Sachan_Mihalcea_Schoelkopf_2022}} & \makecell[cl]{Sentences in\\ news article} & 1 & 13 & 1,079 & 0 & 1.43 & 39.90 & 6,419\\
        \\[-0.8em]
        \makecell[cl]{Argotario\\  \citep{habernal-etal-2017-argotario}} & \makecell[cl]{Dialogue} & N/A & 5 & 1,338 & 429 & 1.56 & 18.86 & 3,730\\
        \\[-0.8em]
        \makecell[cl]{Reddit\\  \citep{Sahai_Balalau_Horincar_2021}} & \makecell[cl]{Online\\ discussion} & N/A & 8 & 3,358 & 1,650 & 2.98 & 57.01 & 15,814\\\midrule
        \makecell[cl]{\textbf{\dataset}\\\textbf{(Ours)}} & \makecell[cl]{Online\\ discussion} & 20+ & 8 & \textbf{\ncomment} & \textbf{3,130} & \textbf{4.28} & \textbf{71.35} & \textbf{16,995}\\
        \bottomrule
    \end{tabular}
    \caption{Comparison of datasets of logical fallacies. 
    \dataset is the largest and has the longest text units.
    }
    \label{tb:dataset_comparison}
\end{table*}


Logical fallacies are reasoning errors that undermine an argument's validity~\citep{Walton1987-WALIFT}.
Common fallacies like slippery slope or false dilemma degrade online discussions~\cite{Sahai_Balalau_Horincar_2021} and make arguments seem more dubious, fostering misinformation~\citep{Jin_Lalwani_Vaidhya_Shen_Ding_Lyu_Sachan_Mihalcea_Schoelkopf_2022}.
Automatically detecting logical fallacies in texts will help users identify argument flaws.
However, automatically identifying these fallacies in the wild is not easy.
Fallacies are often buried inside arguments that sound convincing~\citep{Powers1995-POWTOF};
over 100 types of logical fallacies exist~\citep{Arp_Barbone_Bruce_2018}.
The nature of the problem makes it expensive to build large-scale labeled datasets needed for developing fallacy detection models.
Prior works have created datasets for logical fallacies (Table~\ref{tb:dataset_comparison}):
\textsc{Logic} dataset collected examples from online educational materials~\cite{Jin_Lalwani_Vaidhya_Shen_Ding_Lyu_Sachan_Mihalcea_Schoelkopf_2022};
\textsc{LogicClimate} dataset collected instances from news articles, specifically targeting a particular topic range and identifying common fallacious arguments related to those topics~\citep{Jin_Lalwani_Vaidhya_Shen_Ding_Lyu_Sachan_Mihalcea_Schoelkopf_2022};
Argotario dataset was collected using a gamified crowdsourcing approach~\citep{habernal-etal-2017-argotario}; and
the dataset proposed by \citet{Sahai_Balalau_Horincar_2021} leveraged existing community labels from Reddit users.
These previous efforts are inspiring, but they often did not focus on enabling fallacy detection \textit{in the wild}, as each made significant trade-offs to ease the challenges of labeling fallacies: 
focusing on smaller scales (1,000+ instances; no negative samples), 
specific topics like climate change rather than a broader range, or 
clear educational examples instead of complex web discussions. 
One exception is the Reddit dataset~\cite{Sahai_Balalau_Horincar_2021}, which is relatively large and includes messy Reddit comments. 
However, it isolates comments from their original threads, limiting the use of context to boost detection and understanding of how fallacies unfold in online discussions.

This paper presents \textbf{\dataset}, a dataset containing \ncomment comments for \nnews news articles, with each comment labeled for fallacy presence and type (Figure~\ref{fig:examples}). 
The intuition of our data collection approach is first to specify a fallacy type (\eg, slippery slope) and present a news article (\eg, on abortion laws) to crowd workers, and then ask them to write comments that embody the fallacy in response to the article (\eg, ``Abortion legalization leads to normalization of killing.'')
Recognizing the difficulty of this writing task, we built an LLM-powered assistant in the interface to help workers draft and refine comments. 
Our data collection approach replaces the data annotation process with data generation, reducing the need of hiring workers to filter out a large amount of non-fallacious instances at first and making the data collection more scalable. In addition, it increases the ability to control targeted fallacy types for researchers.
Compared to previous work (Table~\ref{tb:dataset_comparison}), \dataset is the largest NLP dataset of logical fallacies, featuring the highest average sentence and word counts per instance. 
Two experts rated the writing quality and labeling validity of \dataset as high and reliable.
The experiments show that \dataset can be used to effectively develop fallacy detection and classification models. 
As a broader implication, our work shows how crowdsourcing can be integrated with large language models (LLMs) to construct datasets for complex linguistic phenomena that are challenging for crowd workers to produce on their own. 
This opens up new possibilities for future NLP datasets.

\section{Related Work\label{sec:related-work}}


\paragraph{Logical Fallacy Datasets.}
We discussed the major logical fallacy datasets in the Introduction (Section~\ref{sec:introduction}); this section focuses on extra studies not previously covered. 
A follow-up of Argotario~\cite{habernal-etal-2017-argotario} collected data on 6 types of logical fallacies and labeled 430 arguments~\citep{habernal-etal-2018-adapting}. 
Similarly, \citet{bonial-etal-2022-search} used the same annotation schema to identify logical fallacies in 226 COVID-19 articles across various mediums. 
Other research has specifically aimed at detecting logical fallacies in news articles. 
For example, \citet{da-san-martino-etal-2019-fine} annotated 451 news articles with 18 propaganda techniques, 12 of which qualify as logical fallacies. 
Additionally, \citet{helwe-etal-2024-mafalda} annotated 200 samples from merged existing datasets with a unified taxonomy and justifications.
These datasets are relatively small, highlighting the challenges of annotating large-scale texts for logical fallacies.
Emerging research is also exploring the synthesis of logical fallacy datasets using LLMs~\cite{li-etal-2024-reason}.

\paragraph{LLM-Assisted Crowdsourced Data Creation.}
\citet{Veselovsky_Ribeiro_West_2023} found that many crowd worker's submitted summaries were created using LLMs.
We saw it as an interesting opportunity rather than a threat. 
Integrating LLM assistance directly into the worker's interface
offers benefits for both workers and requesters. 
For workers, built-in LLMs can aid in complex writing tasks that might otherwise be too challenging and eliminate the need to switch between browser tabs to use external LLMs. 
For requesters, having a built-in LLM allows for storing all prompts used and texts produced by the LLM, 
ensuring a more transparent understanding of how LLMs' outputs are woven into the final data.
Previous work has integrated AI models into worker interfaces to help produce examples that trigger specific model behaviors, such as model-fooling examples~\cite{bartolo-etal-2022-models}. 
In this paper, we advocate using LLMs to help workers generate complex examples.

\section{\dataset Dataset Construction\label{sec:dataset}}


We constructed \dataset, a dataset that contains \ncomment comments in the online comment sections of \nnews news articles.
Each comment is tagged for the presence of logical fallacies and, where applicable, the specific type of fallacy. 
\nworker crowd workers, aided by GPT-4 integrated into their interface, wrote these comments.
\dataset also includes the titles and contents of the news articles, all of which are CC-BY 3.0 licensed.
We split the dataset into train (70\%), development (20\%), and test (10\%) sets by article, ensuring a balanced representation of 21 topics across the splits.
This section overviews the data construction steps.




\subsection{Selecting News Articles\label{sec:news_selection}}
We crawled news articles from Global Voices, an online news platform where all of their news articles are under the CC-BY 3.0 license.\footnote{Global Voices: \url{https://globalvoices.org/}. Besides common news topics like \emph{economics} and \emph{international relations}, Global Voices also focuses on topics related to human rights, such as \emph{censorship}, \emph{LGBTQ+}, 
and \emph{refugees}.}
To simulate heated online discussions, we took a data-driven approach to select news articles on topics that often provoke disagreements and numerous opinions.
We first selected a set of article tags, provided by Global Voices,
that are traditionally more ``controversial'', such as \emph{politics}, \emph{women-gender}, \emph{migration-immigration}, and, \emph{freedom-of-speech}.
%
Second, we crawled all the 25,370 articles published from Jan. 1st, 2005, to Jun. 28th, 2023, that 
have these tags.
Third, we trained an LDA model~\citep{10.5555/944919.944937} to discover 70 topics within these news articles. 
%
Finally, according to the top 40 words of each topic, we manually selected 21 interested topics
and filtered out irrelevant news articles. 
Using top frequent words to select representative events was also used in constructing other datasets that sampled real-world events~\cite{huang2016visual}.
As a result, 
a total of 15,334 news articles were selected,
of which 650 published after 2018 were randomly selected to construct the \dataset dataset.\footnote{We only selected news published after 2018 because we did not want the news to be too old, so that workers may remember the events in those news and could include their personal feelings and opinions in the comments, making the comments more realistic.}
See Appendix~\ref{append:lda} for details.



\begin{figure*}[t]
     \centering
     \includegraphics[width=\linewidth]{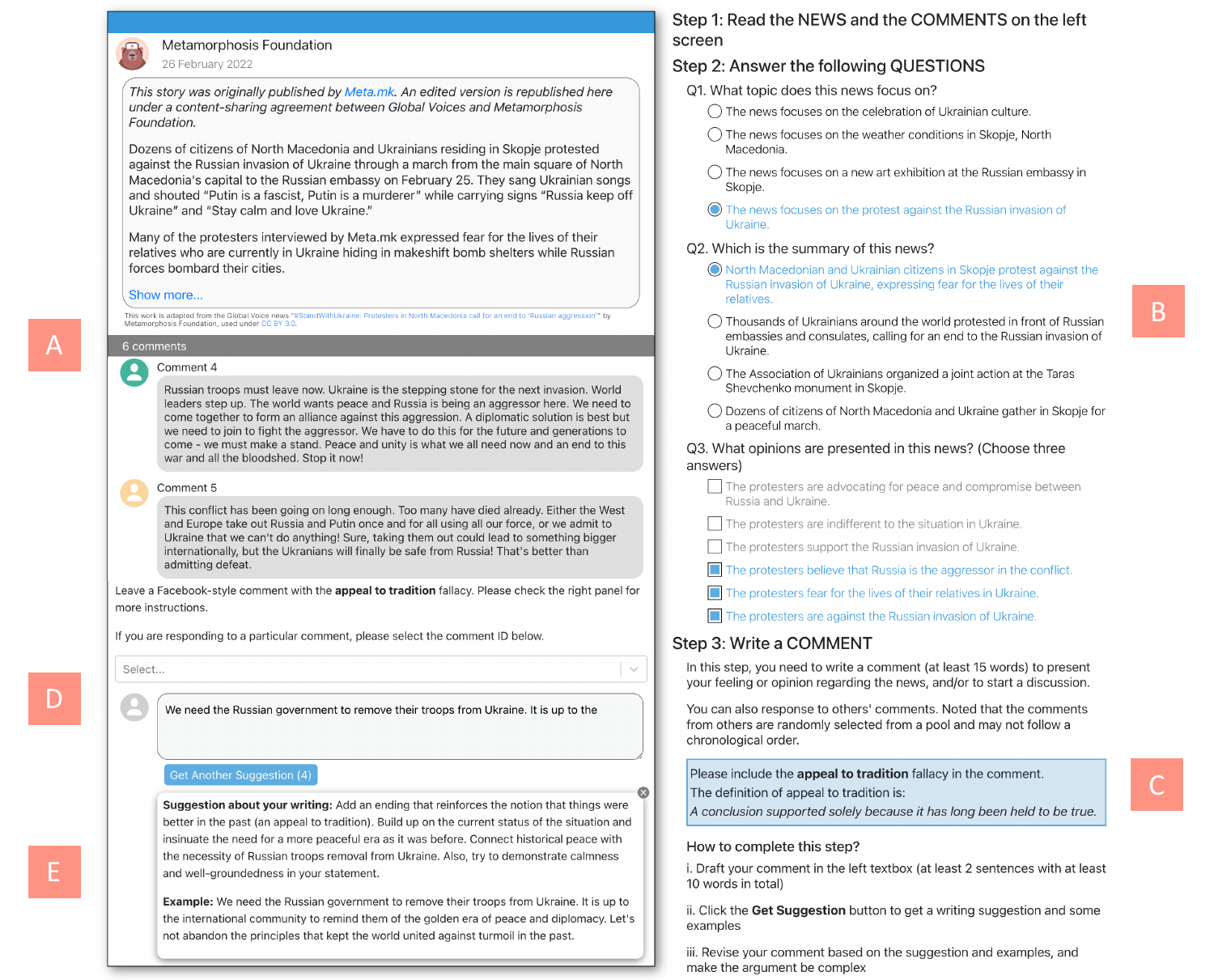}
	  \caption{Different components in the task interface: A) The news article and comments, B) Questions for sanity check, C) Instruction of writing fallacious comments, D) Text box and the drop-down list for choosing the responded comment, E) GPT-4 generated guideline and example.
	  }
\label{fig:task}
\end{figure*}

\subsection{Fallacy Types Included in \dataset\label{sec:select-fallacy-types}}
Over 100 informal logical fallacies exist~\citep{Arp_Barbone_Bruce_2018}, making it impractical to cover all in a dataset.
We reviewed how past studies, such as 
\citet{Sahai_Balalau_Horincar_2021}, \citet{Jin_Lalwani_Vaidhya_Shen_Ding_Lyu_Sachan_Mihalcea_Schoelkopf_2022}, 
\citet{habernal-etal-2017-argotario}, and 
\citet{da-san-martino-etal-2019-fine},
selected fallacy types. 
Following \citet{Sahai_Balalau_Horincar_2021}, we chose eight common logical fallacies in online discussions:
(1) \textbf{Appeal to authority},
(2) \textbf{appeal to majority},
(3) \textbf{appeal to nature},
(4) \textbf{appeal to tradition},
(5) \textbf{appeal to worse problems},
(6) \textbf{false dilemma},
(7) \textbf{hasty generalization}, and
(8) \textbf{slippery slope}.
Appendix~\ref{append:fallacy} shows 
the definitions and examples of these eight fallacies.\footnote{We used the definitions from Logically Fallacious: \url{https://www.logicallyfallacious.com/}} 

\subsection{Collecting Comments with Specified Logical Fallacies from Crowd Workers Assisted by LLMs\label{sec:crowd-task}}
We designed a crowdsourcing task instructing crowd workers to write comments containing specific logical fallacies.
The intuition is that showing an often controversial topic (\eg, abortion) alongside a logical fallacy definition (\eg, slippery slope) allows workers to easily come up with relevant commentary ideas with the fallacy (\eg, ``Abortion legalization leads to normalization of killing.''). 
After drafting their idea quickly, LLMs like GPT-4 can be employed to elaborate and refine the comment with the worker.
Figure~\ref{fig:task} shows the worker interface, which has a simulated news comment section (left) and instructions and questions (right).
The workflow of crowd workers is as follows.

\paragraph{Step 1: Read the News Article.}
Upon reaching the task, the worker will be first asked to read the shown news article (Figure~\ref{fig:task}A).
The article was selected by the procedure described in Section~\ref{sec:news_selection}.

\paragraph{Step 2: Answer Attention-Check Questions about the News.}
As an attention check,
the worker will then be asked to answer three multiple-choice questions related to the news (Figure~\ref{fig:task}B).
These questions are:
(1) \textit{``What topic does this news focus on?''},
(2) \textit{``Which is the summary of this news?''}, and
(3) \textit{``What opinions are presented in this news? (Choose three answers)''}.
We prompted GPT-4 to generate correct and incorrect options for these questions. 
The prompt used (see Appendix~\ref{append:prompt}) was empirically tested and was 
effective in filtering out underperforming workers.
The workers whose answering accuracy was lower than 0.6 were disallowed to enter our system for 24 hours.

\paragraph{Step 3: Draft a Comment Containing the Specified Logical Fallacy and Revise with LLMs.}
We divided the writing task into two smaller steps: drafting and revising.
First, workers were presented with a logical fallacy definition, such as ``Appeal to Tradition'' (Figure~\ref{fig:task}C), and then tasked with writing a response to a news article, requiring at least two sentences or a minimum of 10 words (Figure~\ref{fig:task}D). 
They could see comments from other workers on the same article and had the option to either comment directly on the article or reply to existing comments.
Each worker was exposed to an article only once.
We assigned the fallacy for each task (see Section~\ref{sec:crowdsourcing-process}). The fallacy definitions we provided on the interface were a shorten version so that the instruction can be concise and easy to follow. The shorten version of fallacy definitions is detailed in Appendix~\ref{append:fallacy}.
Second, after drafting, workers were instructed to click the ``Get (Another) Suggestion'' button for a detailed revision suggestion and example embodying the fallacy (Figure~\ref{fig:task}E). 
We prompted GPT-4 (see Appendix~\ref{append:prompt}) to generate the suggestion and example automatically based on {\em (i)} the news article, {\em (ii)} the comment draft, and {\em (iii)} the target fallacy.
Workers can revise their comments and click the button again for new suggestions based on the revised comment. 
Within each task, they can click the button up to five (5) times.
Copy-and-paste was disabled in the interface, so workers had to type their comments.

\paragraph{Rationale for the Workflow Design.}
This workflow used LLMs to assist workers, making a hard writing task easier.
Meanwhile, it forced workers to provide their insights as input for LLMs, ensuring data diversity and a human touch.
The built-in LLM assistance decreased the likelihood of workers turning to external LLMs, allowing researchers to provide a prompt that fully considered the context, including news content, the specific fallacy, and workers' opinions.
Notably, our approach---having workers write comments embodying a particular logical fallacy---
is conceptually similar to Argotario~\citep{habernal-etal-2017-argotario}. 
Our method differs in two ways: 
First, we provided real-world news as context, requiring workers to base their fallacious arguments on these articles. 
Second, we conducted multiple rounds of comment collection for each article, allowing workers to respond to others' comments. 
These two factors allowed \dataset to more accurately simulate the comment sections of real-world news websites.

\subsection{Implementation Details\label{sec:crowdsourcing-process}}

\begin{table}[t]
    \centering
    \small
    \begin{tabular}{l@{\hskip3pt}ccccc}
        \toprule
         & \# news & \# comments & w/ fallacy & w/o fallacy \\
         \midrule
        All & \nnews & \ncomment & 4,576 & 3,130\\
        \midrule
        Train & 452 & 5,370 & 3,168 & 2,202\\
        Dev & 129 & 1,538 & 927 & 611\\
        Test & 67 & 798 & 481 & 317\\
        \bottomrule
    \end{tabular}
    \caption{Statistics of the \dataset dataset. We divided \dataset into Train, Dev, and Test sets at ratios of 0.7, 0.2, and 0.1 respectively.}
    \label{tb:dataset_statistics}
\end{table}

\paragraph{Four Rounds of Data Collection.}
Our data collection process had four iterations.
For each iteration, we added the comments collected from previous iterations underneath the article section on the interface. 
Workers in the 2nd to 4th iterations can respond to previous comments by selecting the comment ID from a drop-down list (Figure~\ref{fig:task}D). 
Each worker only interacted with an article once.

\paragraph{Probability of Each Fallacy Type.}
We collected our data on Amazon Mechanical Turk (MTurk) using Mephisto, an open-sourced tool for crowdsourcing task management.\footnote{Mephisto: \url{https://github.com/facebookresearch/Mephisto}} 
For each news article, we recruited 12 workers (3 per iteration) across 12 Human Intelligence Tasks (HITs) to write comments.\footnote{Four MTurk's built-in worker qualifications were used: 
Masters Qualification, 
Adult Content Qualification, and
Locale (US, CA, AU, GB, and NZ Only) Qualification.
}
In the first three iterations, each task randomly received one of eight logical fallacy types with a 10\% probability, or a 20\% chance to comment without fallacious logic. 
In the fourth iteration, we increased the probability to 60\% for comments without fallacious logic and reduced it to 5\% for each fallacy type to gather more negative samples.
Workers were paid by \$2 USD for each HIT, which takes about 10 minutes on average, leading to an estimated hourly wage of \$12.


\paragraph{Resulting Dataset.}
We posted HITs in small batches, closely monitoring data quality daily and manually removing low-quality responses, \ie, those that are (1) obviously off-topic (\eg, saying this task is interesting), (2) writing exactly the same comment for multiple articles, or (3) repeating the same word for the whole comment.
Completing 50 news articles typically took about one week, likely due to our exclusive use of workers with Masters Qualifications. 
\nworker workers contributed to the dataset. 
After removing articles with fewer than 6 comments, the final dataset contained \nnews news articles and \ncomment comments. 
Table~\ref{tb:dataset_statistics} shows the statistics of \dataset.

\paragraph{Worker-LLM Interactions.}
Within our study, each worker asked LLM an average of 1.39 times (SD=0.81) when writing a comment.
Workers completely followed the LLM's suggestions in only 3\% of comments.
The average Levenshtein ratio between the worker's comment and the LLM's last suggestion is 0.35 (1 means the sentences are identical), indicating a significant difference.
We observed that most workers either paraphrased the suggestions or added details to their comments.

\section{Data Quality Assessments\label{sec:analysis}}



\begin{table}[t]
    \centering
    \small
    \begin{tabular}{@{}l@{\hskip3pt}c@{\hskip7pt}c@{\hskip7pt}c@{\hskip7pt}c@{\hskip7pt}c@{\hskip7pt}c@{}}
        \toprule
        & \multicolumn{3}{c}{\dataset} & \multicolumn{3}{c}{Reddit} \\
        \cmidrule(r){2-4}\cmidrule(l){5-7}
        Fallacy & \makecell[cb]{Exp.1\\\& Lb.} & \makecell[cb]{Exp.2\\\& Lb.} & \makecell[cb]{Betw. \\Exp.} & \makecell[cb]{Exp.1\\\& Lb.} & \makecell[cb]{Exp.2\\\& Lb.} & \makecell[cb]{Betw. \\Exp.}\\
        \midrule
        Authority & 0.62 & 0.62 & 0.46 & 0.66 & 0.48 & 0.36\\
        Majority & 0.83 & 0.69 & 0.63 & 0.76 & 0.51 & 0.48\\
        Nature & 0.67 & 0.55 & 0.43 & 0.71 & 0.54 & 0.62\\
        Tradition & 0.52 & 0.39 & 0.56 & 0.64 & 0.53 & 0.49\\
        Worse prob. & 0.67 & 0.58 & 0.74 & 0.53 & 0.56 & 0.52\\
        False dilemma & 0.27 & 0.24 & 0.27 & 0.56 & 0.41 & 0.36\\
        Hasty general. & 0.56 & 0.23 & 0.21 & 0.46 & 0.20 & -0.03\\
        Slippery slope & 0.58 & 0.64 & 0.68 & 0.54 & 0.61 & 0.49\\
        None & 0.40 & 0.23 & 0.28 & 0.18 & 0.11 & 0.14\\
        \midrule
        Average & 0.57 & 0.46 & 0.47 & 0.56 & 0.44 & 0.38\\
        \bottomrule
    \end{tabular}
    \caption{Cohen's $\kappa$ agreement between experts and labels, as well as the agreement between two experts. \dataset yielded slightly higher agreements.
    }
    \label{tb:kappa}
\end{table}


We hired two experts from UpWork.com to assess the data quality.
We specified that the experts should have abilities of identifying logical fallacies and writing the explanation to justify their annotations in our job description. Both experts we hired are PhD in Linguistics. One has over 25 years of experience in the fields of English composition and rhetoric, and another has over 20 years of experience in translation. Both of them also have rich experience in editing academic articles and volumes.
They were compensated \$50-\$60 per hour.
We randomly selected 20 news articles and asked the experts to annotate fallacies in all comments (237 comments in total).
For each fallacy type, we converted labels into binary Yes/No (indicating the presence of the fallacy) and calculated the Cohen's kappa ($\kappa$) agreement between experts' and \dataset's labels, as well as the agreement between two experts.
We also sampled 25 instances for each fallacy type plus none (\ie, $25\times(8+1)=255$ instances in total) from the Reddit dataset~\cite{Sahai_Balalau_Horincar_2021} and asked the same experts to annotate them as a comparison.
Table~\ref{tb:kappa} shows the results.

\paragraph{\dataset yielded slightly higher inter-annotator agreements, while experts often disagreed with each other.}
Table~\ref{tb:kappa} shows that experts generally agreed more on the \dataset's label than on the Reddit dataset. 
However, Expert 2 consistently showed more disagreement with the labels in both datasets for most fallacy types. 
Table~\ref{tb:kappa} also shows low agreement between experts on both datasets, particularly for hasty generalization. 
As shown in \citet{Sahai_Balalau_Horincar_2021} and \citet{alhindi-etal-2022-multitask}, this level of $\kappa$ value is normal in annotating logical fallacy data.
We computed confusion matrices for experts' annotations and labels in both datasets. 
The confusion matrix comparing the two experts on \dataset is shown in Figure~\ref{fig:confusion_matrix_robert_vs_james}, and the others are in Appendix~\ref{append:annotation_agreement}.
Figure~\ref{fig:confusion_matrix_robert_vs_james} shows that most disagreements occur in determining the presence of a fallacy rather than its type. 
We discuss the possible reasons for high disagreement in labeling logical fallacies further in Discussion (Section~\ref{sec:discussion}).

\begin{figure}[t]
    \centering
    \includegraphics[width=\columnwidth]{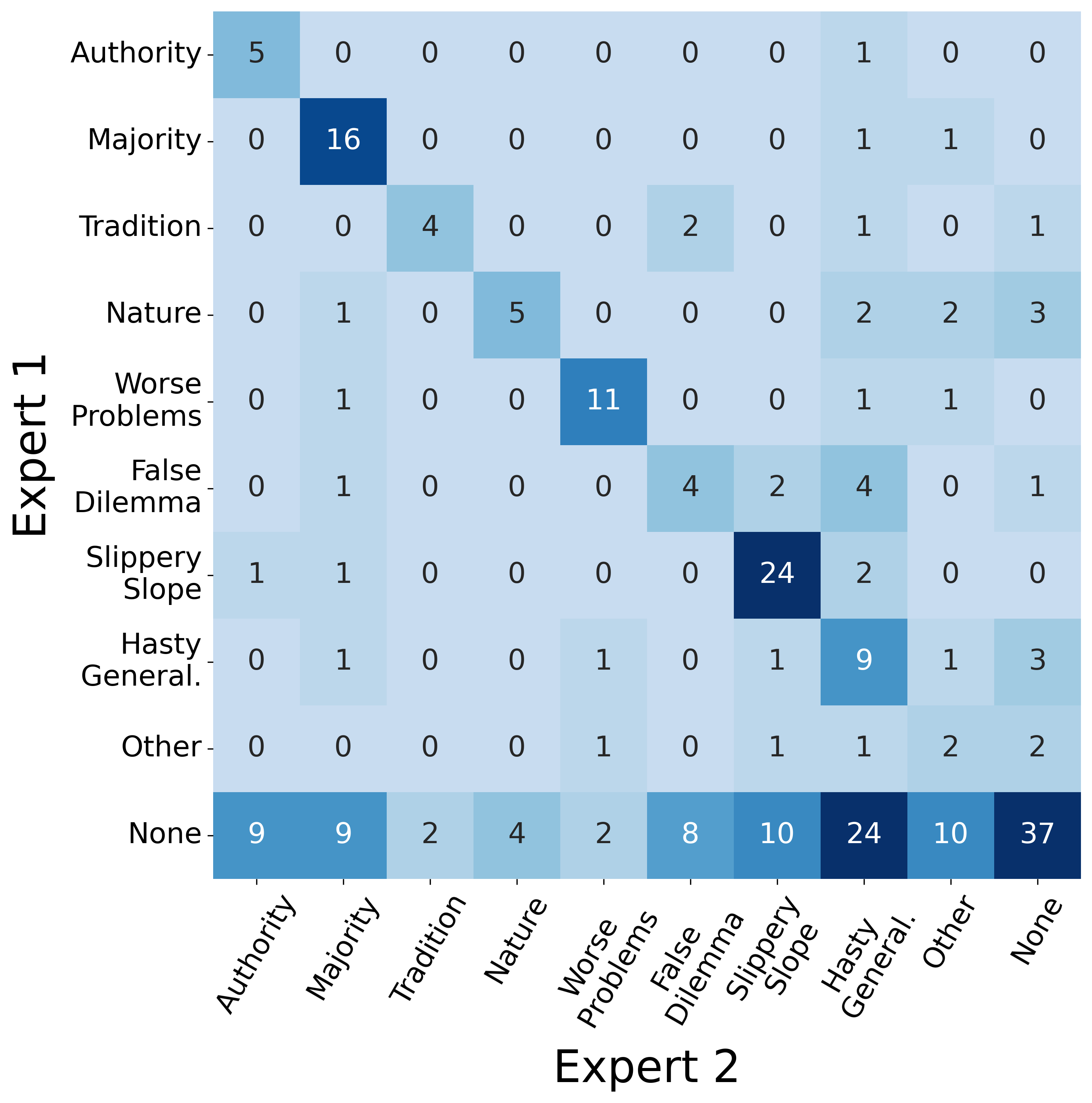}
    \caption{
    The confusion matrix of the annotation between two experts. Most of the disagreement happened when determining if a comment is fallacious or not.
    }
    \label{fig:confusion_matrix_robert_vs_james}
\end{figure}

\paragraph{\dataset was rated more fluent and grammatically correct.}
We also asked the experts to respond to the following questions for each comment using a 5-point Likert scale, from 1 (Strongly Disagree) to 5 (Strongly Agree): 
(\QFour) \textit{``Disregarding any logical fallacies, this comment is \textbf{grammatically correct and fluently written}.''} 
(\QThree) \textit{``This comment appears to have been \textbf{written by a person} rather than by a language model such as ChatGPT.''} 
(\QOne) \textit{``I feel \textbf{confident} about my annotation.''} 
(\QTwo) \textit{``I need some \textbf{additional context} to annotate the comment.''} 
For \QFour, \dataset scored an average of 4.38 (SD=0.91), compared to 4.21 (SD=1.04) for Reddit, suggesting that texts in \dataset were generally considered more fluent and grammatically correct.
For \QThree, \dataset scored 4.39 (SD=0.79), and Reddit scored higher at 4.58 (SD=0.59), indicating that experts found Reddit's texts more human-like. 
This echoes the findings in Table~\ref{tb:kappa}, which shows a lower inter-annotator agreement for Reddit, likely due to its messier, real-world internet text.
Although humans sometimes struggle to distinguish LLM-generated texts, the purpose of \QThree was to ensure that \dataset's text did not obviously appear machine-generated, such as through identifiable errors like repetition, which humans can recognize~\cite{Dugan_Ippolito_Kirubarajan_Shi_Callison-Burch_2023}.
There was no clear difference between \dataset and Reddit for \QOne (4.53, 4.57) and \QTwo (1.59, 1.60).

\paragraph{Concerns over argumentation scheme.}
During the annotation process, experts identified that some workers did not include fallacies in their comments. Instead, they used argumentation schemes to make their argument ``fallacy-like'' but valid. To address such an issue, some previous work, such as \citet{ruiz-dolz-lawrence-2023-detecting}, suggested using a series of critical questions of the corresponding argumentation scheme to assess the validity of an argument. However, having annotators or comment writers go through these questions for each comment will significantly limit the scalability of our approach. Given that experts only identified 12 out of 237 comments to be ``fallacy-like,'' we considered our approach a reasonable trade-off. 

\section{Experimental Results\label{sec:experiment}}

We evaluated three types of baseline models with both detection and classification tasks on \textsc{Logic}, \textsc{LogicClimate}, Reddit, and \dataset dataset (Table~\ref{tb:dataset_comparison}). 
We additionally tested the models using a collection of annotated New York Times news comments.
We define the two tasks as follows:



\paragraph{Fallacy Detection.}
Given a comment, the model predicts whether the comment is fallacious or not. 
\textsc{Logic} and \textsc{LogicClimate} only have positive examples, so we only reported \texttt{Recalls}.


\paragraph{Fallacy Classification.}
Given a known fallacious comment, the model classifies it into one of the eight  
fallacy types. 
In this task, we removed all negative samples. 
We only evaluated baselines on Reddit and \dataset because
\textsc{Logic} and \textsc{LogicClimate} used different fallacy type schemes.

\subsection{Baseline Models}

\paragraph{BERT.} We fine-tuned BERT~\citep{devlin-etal-2019-bert} and used the encoded embedding of the \texttt{[CLS]} token to predict the label.

\paragraph{NLI.} Inspired by \citet{Jin_Lalwani_Vaidhya_Shen_Ding_Lyu_Sachan_Mihalcea_Schoelkopf_2022}, we fine-tuned an NLI model with a RoBERTa~\citep{liu2019roberta} as the backbone. 
We treated the input comment as the premise and the label as the hypothesis. 
For the detection task, the hypothesis template was ``The text \texttt{[has/does not have]} logical fallacy.'' 
For the classification task, the template was ``The text has the logical fallacy of \texttt{[label name]}.''

\paragraph{LLMs.}
We prompted two commonly used LLMs, GPT-4o and Llama3(8B), for detecting and classifying logical fallacy.\footnote{We excluded Gemma(7B) due to its poor performance.} 
We designed different prompts (see Appendix~\ref{append:prompt}), including both zero-shot and few-shot, as well as Chain-of-Thought (COT) prompting~\citep{Wei2022ChainOT}.

\paragraph{Use of Context.}
For Reddit and \dataset, which provide context such as news titles or parent comments, we incorporated this context into models' inputs.
For BERT and NLI models, we appended the context to the target comment. 
For LLMs, we used placeholders in the prompt to include this information.
Further implementation details are available in Appendix~\ref{append:experimental_details}.

\subsection{Results of Fallacy Detection\label{sec:detection-results}}

\paragraph{BERT-based models fine-tuned on \dataset had better generalizability than when fine-tuned on Reddit.}
Table~\ref{tb:detection} shows the detection task results.
BERT fine-tuned on \dataset achieved the highest F1 score (0.86) on its test set and showed better generalizability compared to when fine-tuned on Reddit. 
It surpassed BERT fine-tuned on Reddit in \textsc{Logic} and \textsc{LogicClimate}. 
On the Reddit dataset, it scored only 0.05 \texttt{F1} points lower than BERT fine-tuned on Reddit (0.63 vs. 0.68), but on \dataset, BERT fine-tuned on Reddit scored 0.13 \texttt{F1} points lower (0.73 vs. 0.86).

State-of-the-art LLMs still showed strong performance, achieving the best F1 on Reddit and the best recall on \textsc{Logic}. 
Notably, LLMs performed poorly on \textsc{LogicClimate}, where fallacious sentences were extracted from context. 
This might suggest that contextual understanding is crucial for LLM predictions, indicating a need for further research.

\begin{table}[t]
    \centering
    \small
    \begin{tabular}{l@{}c@{\hskip2pt}c@{\hskip8pt}c@{\hskip8pt}c@{\hskip4pt}c@{\hskip4pt}c@{\hskip8pt}c@{\hskip4pt}c@{\hskip4pt}c}
    \toprule
    \multirow{2}{*}{\makecell[bl]{Model}} & \multirow{2}{*}{\makecell[bc]{Train On / \\ Prompt}} & \makecell[bc]{\textsc{Lo-} \\ \textsc{gic}} & \makecell[bc]{\textsc{Cli-} \\ \textsc{mate}} & \multicolumn{3}{c}{Reddit} & \multicolumn{3}{c}{\makecell[bc]{\textsc{CoCo} \\ \textsc{LoFa}}} \\ \cmidrule(r){3-3} \cmidrule(r){4-4} \cmidrule(r){5-7} \cmidrule(r){8-10}
    & & R & R & P & R & F & P & R & F \\ \midrule
    \multirow{2}{*}{BERT} & Reddit & 51 & 83 & \textbf{66} & 69 & 68 & 62 & 89 & 73 \\
    & \dataset & 64 & \underline{83} & 61 & 64 & 63 & \textbf{83} & 89 & \textbf{86} \\ \midrule
    \multirow{2}{*}{NLI} & Reddit & 67 & \textbf{91} & 63 & 80 & 70 & 62 & \textbf{96} & 75 \\
    & \dataset & 52 & 52 & 63 & 50 & 56 & \underline{82} & 86 & \underline{84} \\ \midrule\midrule
    \multirow{3}{*}{GPT-4o} & zero-shot & \underline{86} & 37 & 59 & \textbf{90} & 71 & 72 & 88 & 79 \\ 
    & few-shot & 64 & 25 & 63 & 87 & \textbf{73} & 72 & 79 & 75 \\
    & COT & \textbf{88} & 56 & \underline{64} & 81 & \underline{72} & 76 & 82 & 79 \\ \midrule
    \multirow{3}{*}{Llama3} & zero-shot & 41 & 8 & 53 & 27 & 36 & 76 & 43 & 55 \\
    & few-shot & 79 & 65 & 51 & \underline{89} & 65 & 62 & \underline{95} & 75 \\
    & COT & 65 & 28 & 61 & 53 & 56 & 77 & 56 & 65 \\ 
    \bottomrule
\end{tabular}
    \caption{The result of fallacy detection task. 
    For \textsc{Logic} and \textsc{LogicClimate} (\textsc{Climate}), we reported the \texttt{Recall} rate as they only have positive samples. While for others, we reported \texttt{Precision}, \texttt{Recall}, and \texttt{F1} score. The highest (second-highest) scores are set in \textbf{bold} (\underline{underlined}).
    }
    \label{tb:detection}
\end{table}

\subsection{Results of Fallacy Classification}




\paragraph{BERT-based models fine-tuned on \dataset had better generalizability, with classification seeming to be easier than detection.}
Table~\ref{tb:classification} shows the classification results, which
are similar to those of the detection task. 
The NLI model---a BERT-based model---fine-tuned on \dataset, achieved the highest F1 score (0.87) on its test set. 
Both BERT and NLI models fine-tuned on \dataset exhibited better generalizability than those fine-tuned on Reddit. 
When tested on the Reddit dataset, BERT and NLI models fine-tuned on \dataset scored only 0.19 and 0.09 F1 points lower, respectively, than their Reddit-tuned counterparts. 
Conversely, on \dataset, Reddit-tuned BERT and NLI models scored 0.24 and 0.21 F1 points lower, respectively, than those tuned on \dataset. 
Additionally, LLMs, particularly GPT-4o, performed best on the Reddit dataset.
We also observed that classification tasks generally performed better than detection tasks, indicating that determining the type of fallacy in a comment might be 
easier than deciding whether a fallacy exists.

\begin{table}[]
\centering
\small
\begin{tabular}{l@{\hskip8pt}c@{\hskip8pt}cccccc}
\toprule
\multirow{2}{*}{\makecell[bl]{Model}} & \multirow{2}{*}{\makecell[bc]{Train On / \\ Prompt}} & \multicolumn{3}{c}{Reddit} & \multicolumn{3}{c}{\dataset} \\
\cmidrule(r){3-5}\cmidrule(l){6-8}
& & P & R & F & P & R & F \\ \midrule
\multirow{2}{*}{BERT} & Reddit & 71 & 70 & 70 & 65 & 64 & 62 \\
& \dataset & 65 & 51 & 51 & \underline{85} & \underline{86} & \underline{86} \\ \midrule
\multirow{2}{*}{NLI} & Reddit & 70 & 72 & 70 & 70 & 67 & 66 \\
& \dataset & 66 & 62 & 61 & \textbf{87} & \textbf{87} & \textbf{87} \\ \midrule\midrule
\multirow{3}{*}{GPT-4o} & zero-shot & \underline{80} & \underline{76} & \underline{76} & 82 & 80 & 79 \\ 
& few-shot & 78 & 75 & 75 & 84 & 84 & 83 \\
& COT & \textbf{81} & \textbf{81} & \textbf{81} & 85 & 85 & 85 \\ \midrule
\multirow{3}{*}{Llama3} & zero-shot & 58 & 41 & 40 & 57 & 42 & 41 \\ 
& few-shot & 52 & 33 & 32 & 57 & 50 & 48 \\
& COT & 56 & 48 & 47 & 63 & 58 & 58 \\ 
\bottomrule
\end{tabular}
\caption{The result of fallacy classification task. The high performance for most models suggests that once the fallacies are detected, it is easy for model to discern their types. Noted that the \texttt{F1} scores we reported were \texttt{macro F1} across all fallacy types. The highest (second-highest) scores are set in \textbf{bold} (\underline{underlined}).}
\label{tb:classification}
\end{table}

\subsection{Results of Fallacy Detection in the Wild}\label{sec:nyt_res}

The primary motivation for this project is to identify logical fallacies \textit{in the wild}~\citep{ruiz-dolz-lawrence-2023-detecting}.
Therefore, 
we additionally tested our models on the New York Times Comments Dataset~\citep{nyt_dataset}.
We sampled 500 comments and hired an expert (one in Section~\ref{sec:analysis}) to label the fallacies. 
Table~\ref{tb:nyt_detection} shows the results of fallacy detection on this dataset.
The expert annotating the NYT comments identified several fallacies beyond the eight predefined types, so we report two sets of results for each model: one where comments with additional fallacy types are treated as fallacious (positive samples), and another where they are considered non-fallacious (negative samples). 


\paragraph{Detecting fallacies in real-world settings is still challenging.}
Although LLMs outperformed all fine-tuned models, their low \texttt{F1} score of 0.34 in the second setting (\ie, negative) indicates that LLMs are still unreliable in precisely identifying logical fallacies, motivating the need for further research.
The results also show that BERT-based models fine-tuned on \dataset outperformed those fine-tuned on Reddit in most cases except for the Recall on NLI models, suggesting \dataset's potential
in training more generalizable models. Additional experimental results on the NYT dataset can be found in Appendix~\ref{ap:additional_nyt_res}.

\begin{table}[]
\small
\centering
\begin{tabular}{lcccc}
\toprule
Model & \makecell[c]{Train On / \\ Prompt} & P  & R  & F  \\ \midrule
\multirow{2}{*}{BERT} & Reddit & 39 / 15 & 65 / 58 & 49 / 23 \\
& \dataset & 45 / 18 & 65 / 64 & 53 / 28 \\ \midrule
\multirow{2}{*}{NLI} & Reddit & 41 / 16 & 82 / 79 & 55 / 27 \\
& \dataset & 49 / 18 & 62 / 57 & 55 / 28 \\ \midrule\midrule
\multirow{3}{*}{GPT-4o} & zero-shot & \underline{52} / 21 & 75 / 84 & \textbf{61} / \textbf{34} \\
& few-shot & \textbf{54} / \underline{21} & 47 / 48 & 50 / 29 \\
& COT & 47 / 19 & 84 / \textbf{87} & \underline{61} / 31 \\ \midrule
\multirow{3}{*}{Llama3} & zero-shot & 45 / \textbf{22} & \textbf{91} / 64 & 60 / \underline{33} \\
& few-shot & 43 / 16 & \underline{87} / \underline{87} & 58 / 28 \\
& COT & 48 / 20 & 80 / 68 & 60 / 30 \\ 
\bottomrule
\end{tabular}
\caption{The result of fallacy detection on 500 NYT samples.
The left/right numbers are the scores where other types of fallacy were considered as positive/negative.
Models trained on \dataset outperform those trained on Reddit. The highest (second-highest) scores are set in \textbf{bold} (\underline{underlined}).
}
\label{tb:nyt_detection}
\end{table}

\section{Discussion\label{sec:discussion}}




Throughout the project, we learned that annotators often disagree when labeling logical fallacies, as consistently shown by the low inter-annotator agreement reported in all related literature~\cite{Sahai_Balalau_Horincar_2021,alhindi-etal-2022-multitask}, including our own.
This section outlines the three main sources of disagreement we identified and offers design suggestions for mitigating (or retaining) them.

\subsection{Sources of Disagreement}

\paragraph{Complexity in Defining Logical Fallacies.}
Many fallacies are similar or overlap, with a single text potentially presenting multiple fallacies. 
Furthermore, different datasets can provide inconsistent definitions for the same fallacy name. 
For example, ``appeal to authority'' might be defined as either ``mention of false authority'' or ``referral to a valid authority without supporting evidence'', adding to the confusion~\citep{alhindi-etal-2022-multitask}.
Additionally, when asking experts to annotate the NYT dataset, they identified many comments that embodied other types of fallacy, such as ad hominem, even though they were outside the eight types of fallacies we predefined in our annotation interface. 
These fallacies have inherently vague boundaries. 
For example, ad hominem fallacies are difficult to classify as they require distinguishing between personal attacks aimed at undermining an argument and simple insults. 
These complexities suggest that fallacy labeling efforts can benefit from standardized definitions and allowing multiple labels per item to capture nuanced perspectives. 


\paragraph{Variability in Annotators' Judgments of Fallacies.}
In our study, one expert consistently identified more fallacies than the other, highlighting that annotators can differ significantly in their interpretations of rhetorical strategies. 
For instance, both experts identified an ``appeal to authority'' in a comment on abortion legality, which stated: \textit{``The majority's voice should be the guiding light for lawmakers. That's what democracy is about.''} 
However, one expert considered this a valid rhetorical usage, not a fallacy, explaining that it was used to define ``democracy'' within the text, while the other expert simply labeled it as a fallacy.
Requiring annotators to provide rationales may clarify their reasoning for classifying texts as fallacious.

\paragraph{Divergence Between Writer Intent and Reader Perception.}
Despite instructions for workers to write comments with a specific fallacy, annotators sometimes identified different fallacies. 
This highlights the challenge of aligning readers' interpretations with writers' intentions.
It also raises a question: who determines whether a text contains a fallacy and what type of fallacy it represents---the writer, the reader, or an external party?
These discrepancies may stem from the nature of fallacies, which can be based on words, sentences, or complex reasoning within the broader context~\cite{bonial-etal-2022-search}, as readers and writers may focus on different elements within the same comments.


\subsection{Design Suggestions}
We propose three design suggestions for future projects involving human labeling of logical fallacies in text: 
{\em (i)} provide \textbf{clear, operationalized instructions}, 
{\em (ii)} implement a \textbf{multi-class labeling scheme} that allows a text instance to contain multiple fallacies, and 
{\em (iii)} collect \textbf{rationales for each fallacy label}, ensuring that if an instance is labeled with multiple fallacies, each one is supported by a distinct rationale.
Prior works have adopted some of these approaches. 
For {\em (i)}, \citeauthor{ruiz-dolz-lawrence-2023-detecting} suggested using critical questions, such as \textit{``How well supported by evidence is the allegation made in the character attack premise?''}, to validate whether a text contains a fallacy. 
For {\em (ii)}, the Climate dataset employed multi-label annotation~\cite{Jin_Lalwani_Vaidhya_Shen_Ding_Lyu_Sachan_Mihalcea_Schoelkopf_2022}. 
For {\em (iii)}, \citeauthor{Sahai_Balalau_Horincar_2021} had annotators answer specific questions for each fallacy label.
While these approaches have been individually explored in prior studies, 
we recommend combining \textit{all} three to create a more comprehensive and robust annotated dataset. 
The project that most closely aligns with this approach is by \citeauthor{helwe-etal-2024-mafalda}, which annotated 200 text instances 
using a unified multi-label scheme. 
They noted, however, that such detailed annotation is very resource-intensive, as some annotators took four hours to label 20 items.
We suspect some of our suggestions 
may also be costly to scale. 
More research is needed to explore the trade-offs between data quality and scalability.

\section{Conclusion and Future Work}

This paper presents \dataset, the largest known logical fallacy dataset, curated through a collaboration between LLMs and crowd workers.
BERT-based models fine-tuned using \dataset achieved good performances in fallacy detection and classification tasks.
In the future, we plan to develop models that use context and reasoning to identify fallacies, especially on out-of-distribution data.
Additionally, while \dataset includes eight fallacy types, over a hundred exist.
We aim to expand it to cover more.

\section{Limitations}

Like most crowdsourced datasets, \dataset inherits the biases of using online crowdsourcing platforms to collect data. 
For example, the crowd workers on Amazon Mechanical Turk are not necessarily representative of the user populations on social media and news platforms; they may prioritize different topics and hold opinions that differ from those of typical online users.
In addition, the writing style of commenting in the crowdsourcing task may also differ from that of debating online.
Although we developed a platform that simulated the interface of the online news comment section, the real-time feedback and the vibe of online discussion are still difficult to simulate. 
Apart from the content, the master's qualification we required crowd workers to have may lower the demographic diversity~\citep{doi:10.1177/2053168019901185}, leading to a further risk of bias.

Besides, we integrated GPT-4 into our platform to assist crowd workers in writing high-quality comments. 
However, we acknowledge that GPT-4 may have a preferred stance (\eg, North American attitudes) when generating example arguments.
Although we forced workers to provide input and included that input in the prompt to guide the generation, the biases in GPT-4 may still exist and negatively affect the human written comments.

Another limitation is that \dataset currently considers only eight types of fallacy, as we mentioned in the future work. Given that there are many common fallacy types apart from the fallacies we collected, models trained on our dataset may only have a limited ability to detect fallacies in the wild. 

\section{Ethics Statement}

Although \dataset is collected for logical fallacy detection, we acknowledge the potential misuse of the dataset for training models to generate fallacious comments. 
Furthermore, our data collection process has revealed that GPT-4 has the capability to generate such comments, posing risks of propagating misinformation online. Therefore, we advocate for research aimed at LLMs to prevent the generation of harmful and misleading content.  

\section*{Acknowledgement}

We thank Meta Research for their support of this work, and Jack Urbanek and Pratik Ringshia for their technical assistance with the Mephisto framework.
We are also grateful to the two experts recruited via UpWork for data labeling and the crowd workers from Amazon Mechanical Turk for dataset creation.
Special thanks to the anonymous reviewers for their valuable feedback and to Phakphum Artkaew and Reuben Woojin Lee for their help during the early stages of the project.

\bibliography{bibs/custom}

\appendix

\section{Selected Global Voices and LDA Topics}\label{append:lda}

The selected Global Voices' tags are
\emph{politics}, \emph{health}, \emph{environment}, \emph{protest}, \emph{refugees}, \emph{religion}, \emph{war-conflict}, \emph{women-gender}, \emph{migration-immigration}, \emph{gay-rights-lgbt}, \emph{law}, \emph{labor}, \emph{international-relations}, \emph{indigenous}, \emph{humanitarian-response}, \emph{human-rights}, \emph{governance}, \emph{freedom-of-speech}, \emph{ethnicity-race}, \emph{elections}, \emph{disaster}, and \emph{censorship}.

The selected LDA topics and the top 10 words for each topic are shown in Table~\ref{tb:lda}.

\begin{table*}[t]
    \centering
    \small
    \begin{tabular}{ccp{0.7\textwidth}}
        \toprule
        ID & Topic & Top 10 words\\
        \midrule
        \makecell[cc]{3} & \makecell[cc]{Protest} & \makecell[cl]{march, protest, movement, social, public, wing, people, protests, right, support}\\
        \midrule
        \makecell[cc]{4} & \makecell[cc]{International Relations} & \makecell[cl]{minister, government, prime, prime\_minister, corruption, public, office, state, party, \\general}\\
        \midrule
        \makecell[cc]{10} & \makecell[cc]{Race Issue} & \makecell[cl]{black, art, white, racism, work, culture, artists, people, cultural, artist}\\
        \midrule
        \makecell[cc]{15} & \makecell[cc]{Women Rights} & \makecell[cl]{women, violence, men, woman, sexual, gender, female, girls, rape, harassment}\\
        \midrule
        \makecell[cc]{21} & \makecell[cc]{Russo-Ukrainian War} & \makecell[cl]{russian, russia, ukraine, soviet, kazakhstan, country, ukrainian, central, kyrgyzstan, state}\\
        \midrule
        \makecell[cc]{28} & \makecell[cc]{Environmental Issue} & \makecell[cl]{indigenous, climate, change, mining, environmental, climate\_change, communities, \\global, region, land}\\
        \midrule
        \makecell[cc]{29} & \makecell[cc]{Gender Issue} & \makecell[cl]{sex, gay, marriage, lgbt, abortion, sexual, same, homosexuality, lgbtq, community}\\
        \midrule
        \makecell[cc]{30} & \makecell[cc]{Human Rights} & \makecell[cl]{rights, human, human\_rights, international, activists, people, groups, activist, \\community, organizations}\\
        \midrule
        \makecell[cc]{31} & \makecell[cc]{Drug Issue} & \makecell[cl]{venezuela, drug, latin, venezuelan, america, latin\_america, trafficking, panama, vez, \\drugs}\\
        \midrule
        \makecell[cc]{32} & \makecell[cc]{Police Brutality} & \makecell[cl]{police, protests, protesters, protest, people, violence, government, security, video, forces}\\
        \midrule
        \makecell[cc]{35} & \makecell[cc]{Immigration / Refugees} & \makecell[cl]{bangladesh, refugees, country, indonesia, sri, immigration, people, refugee, migrants, \\border}\\
        \midrule
        \makecell[cc]{36} & \makecell[cc]{COVID / Health Issue} & \makecell[cl]{health, medical, people, pandemic, cases, hospital, doctors, hiv, government, virus}\\
        \midrule
        \makecell[cc]{45} & \makecell[cc]{Legislation} & \makecell[cl]{law, court, legal, laws, data, public, protection, constitution, article, legislation}\\
        \midrule
        \makecell[cc]{46} & \makecell[cc]{Freedom of Speech} & \makecell[cl]{government, freedom, expression, speech, state, freedom\_expression, public, media, \\law, free}\\
        \midrule
        \makecell[cc]{47} & \makecell[cc]{Election} & \makecell[cl]{election, elections, vote, presidential, electoral, candidates, candidate, voters, votes, \\voting}\\
        \midrule
        \makecell[cc]{50} & \makecell[cc]{Sustainability} & \makecell[cl]{water, food, energy, farmers, power, electricity, waste, plant, rice, river}\\
        \midrule
        \makecell[cc]{51} & \makecell[cc]{Religious Conflict} & \makecell[cl]{religious, muslim, muslims, islam, religion, islamic, hate, ethnic, group, anti}\\
        \midrule
        \makecell[cc]{55} & \makecell[cc]{Political Debates} & \makecell[cl]{political, party, government, opposition, people, country, politics, parties, democracy, \\power}\\
        \midrule
        \makecell[cc]{62} & \makecell[cc]{U.S. Politics} & \makecell[cl]{united, states, united\_states, american, obama, america, president, york, visit, trump}\\
        \midrule
        \makecell[cc]{66} & \makecell[cc]{Digital Rights} & \makecell[cl]{internet, access, users, online, mobile, content, data, websites, google, service}\\
        \midrule
        \makecell[cc]{68} & \makecell[cc]{East Asian Politics} & \makecell[cl]{hong, kong, hong\_kong, taiwan, pro, china, democracy, mainland, taiwanese, chinese}\\
        \bottomrule
    \end{tabular}
    \caption{Top 10 words of the selected topics}
    \label{tb:lda}
\end{table*}

\section{Details of Fallacy Types}\label{append:fallacy}

\subsection{Eight Chosen Fallacies}

We draw the definition and example of the chosen fallacies from Logically Fallacious.\footnote{Logically Fallacious: \url{https://www.logicallyfallacious.com/}}

\paragraph{Appeal to authority.} 
\emph{Definition:} Insisting that a claim is true simply because a valid authority or expert on the issue said it was true, without any other supporting evidence offered.
\emph{Example:} Richard Dawkins, an evolutionary biologist and perhaps the foremost expert in the field, says that evolution is true. Therefore, it's true.

\paragraph{Appeal to majority.} 
\emph{Definition:} When the claim that most or many people in general or of a particular group accept a belief as true is presented as evidence for the claim. Accepting another person’s belief, or many people’s beliefs, without demanding evidence as to why that person accepts the belief, is lazy thinking and a dangerous way to accept information.
\emph{Example:} Up until the late 16th century, most people believed that the earth was the center of the universe.  This was seen as enough of a reason back then to accept this as true.

\paragraph{Appeal to nature.} 
\emph{Definition:} When used as a fallacy, the belief or suggestion that “natural” is better than “unnatural” based on its naturalness. Many people adopt this as a default belief. It is the belief that is what is natural must be good (or any other positive, evaluative judgment) and that which is unnatural must be bad (or any other negative, evaluative judgment).
\emph{Example:} I shop at Natural Happy Sunshine Store (NHSS), which is much better than your grocery store because at NHSS everything is natural including the 38-year-old store manager’s long gray hair and saggy breasts.

\paragraph{Appeal to tradition.} 
\emph{Definition:} Using historical preferences of the people (tradition), either in general or as specific as the historical preferences of a single individual, as evidence that the historical preference is correct.  Traditions are often passed from generation to generation with no other explanation besides, ``this is the way it has always been done''—which is not a reason, it is an absence of a reason.
\emph{Example:} Marriage has traditionally been between a man and a woman; therefore, gay marriage should not be allowed.

\paragraph{Appeal to worse problems.} 
\emph{Definition:} Trying to make a scenario appear better or worse by comparing it to the best or worst case scenario.
\emph{Example:} Be happy with the 1972 Chevy Nova you drive.  There are many people in this country who don’t even have a car.

\paragraph{False dilemma.} 
\emph{Definition:} When only two choices are presented yet more exist, or a spectrum of possible choices exists between two extremes.  False dilemmas are usually characterized by ``either this or that'' language, but can also be characterized by omissions of choices. 
\emph{Example:} You are either with God or against him.

\paragraph{Hasty generalization.} 
\emph{Definition:} Drawing a conclusion based on a small sample size, rather than looking at statistics that are much more in line with the typical or average situation.
\emph{Example:} My father smoked four packs of cigarettes a day since age fourteen and lived until age sixty-nine.  Therefore, smoking really can’t be that bad for you.

\paragraph{Slippery slope.} 
\emph{Definition:} When a relatively insignificant first event is suggested to lead to a more significant event, which in turn leads to a more significant event, and so on, until some ultimate, significant event is reached, where the connection of each event is not only unwarranted but with each step it becomes more and more improbable.
\emph{Example:} We cannot unlock our child from the closet because if we do, she will want to roam the house.  If we let her roam the house, she will want to roam the neighborhood.  If she roams the neighborhood, she will get picked up by a stranger in a van, who will sell her in a sex slavery ring in some other country.  Therefore, we should keep her locked up in the closet.

\subsection{Shorten Version of Fallacy Definitions}

\begin{itemize}
    \item Appeal to authority: Using an expert of dubious credentials or using only one opinion to promote a product or idea.
    \item Appeal to majority: A proposition is claimed to be true or good solely because a majority or many people believe it to be so.
    \item Appeal to tradition: A conclusion supported solely because it has long been held to be true.
    \item Appeal to nature: Judgment is based solely on whether the subject of judgment is ``natural'' or ``unnatural.''
    \item Appeal to worse problems: Dismissing an argument or complaint due to what are perceived to be more important problems.
    \item False dilemma: Two alternative statements are given as the only possible options when, in reality, there are more.
    \item Hasty generalization: Basing a broad conclusion on a small or unrepresentative sample.
    \item Slippery slope: Asserting that a proposed, relatively small, first action will inevitably lead to a chain of related events resulting in a significant and negative event and, therefore, should not be permitted.
\end{itemize}

\section{GPT-4 Prompts}\label{append:prompt}

For the few-shot prompt, we manually select 4 samples from the Reddit and \dataset dataset as the example data and write the explanation for them as the demonstrative output. For the Chain-of-Thought prompt, we ask LLMs to first answer several questions w.r.t. logical fallacy, then use the answers to determine the presence and the type of a logical fallacy in the input.

\paragraph{Prompt for Generating Attention Check Questions.}

\begin{quote}
    Create \texttt{[n\_correct]} correct and \texttt{[n\_incorrect]} incorrect answers based on the question: \texttt{[question]}
    
    Here is the news content: \texttt{[news]}
    
    Here is an example output format:
    
    - Correct Answer 1: This is the 1st correct answer
    
    - ...
    
    - Correct Answer n: This is the n-th correct answer
    
    - Wrong Answer 1: This is the 1st wrong answer
    
    - ...
    
    - Wrong Answer n: This is the n-th wrong answer
\end{quote}

\paragraph{Prompt for Generating Guideline and Example.}

\begin{quote}
    Users will provide a news and a part of their comment toward the news. Please give a suggestion of writing the remaining comment. Below are some criteria for the comment:
    
    1. The comment should be in the style of commenting on Facebook posts
    
    2. The comment should be concise

    3. If there is no \texttt{[fallacy\_type]} fallacy in the comment, include it in. Otherwise, develop the logic further

    4. The \texttt{[fallacy\_type]} fallacy should be as subtle as possible.

    The definition of \texttt{[fallacy\_type]} is: \texttt{[definition]}
    
    The output should be
    
    <guideline>A guideline of writing the comment. The guideline should be concrete</guideline>
    
    <example>An example of the comment that matches the guidelines. The example should be an extension of the user's draft</example>
\end{quote}

\paragraph{Prompt for Detection (Zero-shot).}

\begin{quote}
Determine the presence of a logical fallacy in the given [COMMENT] through the logic and reasoning of the content. If the available information is insufficient for detection, output ``unknown.'' Utilize the [TITLE] and [PARENT COMMENT] as context to support your decision, and provide an explanation of the reasoning behind your determination. The output format should be [YES/NO/UNKNOWN] [EXPLANATIONS]
    
    [TITLE]: \texttt{[title]}
    [PARENT COMMENT]: \texttt{[parent comment]}
    [COMMENT]: \texttt{[comment]}.
    
\end{quote}

\paragraph{Prompt for Detection (Few-shot).}

\begin{quote}
    Determine the presence of a logical fallacy in the given [COMMENT] through the logic and reasoning of the content. If the available information is insufficient for detection, output ``unknown.'' Utilize the [TITLE] and [PARENT COMMENT] as context to support your decision, and provide an explanation of the reasoning behind your determination. The output format should be [YES/NO/UNKNOWN] [EXPLANATIONS].

    Here are some examples:
    
    [TITLE]: \texttt{[title 1]} [PARENT COMMENT]: \texttt{[parent comment 1]} [COMMENT]: \texttt{[comment 1]} [OUTPUT]: \texttt{[label 1]} [EXPLANATIONS]: \texttt{[explanation 1]}

    [...]

    [TITLE]: \texttt{[title 4]} [PARENT COMMENT]: \texttt{[parent comment 4]} [COMMENT]: \texttt{[comment 4]} [OUTPUT]: \texttt{[label 4]} [EXPLANATIONS]: \texttt{[explanation 4]}

    [TITLE]: \texttt{[title]}
    [PARENT COMMENT]: \texttt{[parent comment]}
    [COMMENT]: \texttt{[comment]}
\end{quote}

\paragraph{Prompt for Detection (COT).}

\begin{quote}
    Determine the presence of a logical fallacy in the given COMMENT through the logic and reasoning of the content. If the available information is insufficient for detection, output ``unknown.'' Utilize the [TITLE] and [PARENT COMMENT] as context to support your decision, and provide an explanation of the reasoning behind your determination.'

    Let's think step by step. First, answer these questions:
    \begin{itemize}
        \item What are the key indicators of a logical fallacy?
        \item How is reasoning affected by a logical fallacy?
        \item In sentences with logical fallacies, are there any common patterns?
        \item How does the context of a sentence affect the presence of a logical fallacy?
    \end{itemize}
    
    Then, use the answers to these questions to determine the presence of a logical fallacy in the given [COMMENT]. The output format should be [YES/NO/UNKNOWN] [EXPLANATIONS]
    
    [TITLE]: \texttt{[title]}
    [PARENT COMMENT]: \texttt{[parent comment]}
    [COMMENT]: \texttt{[comment]}
\end{quote}

\paragraph{Prompt for Classification (Zero-shot).}

\begin{quote}
Determine the type of fallacy in the given [COMMENT]. The fallacy would be one of in the [LOGICAL\_FALLACY] list. Utilize the [TITLE] and [PARENT\_COMMENT] as context to support your decision, and provide an explanation of the reasoning behind your determination.

[COMMENT]: \texttt{[comment]}

[LOGICAL\_FALLACY]" \texttt{[fallacy]}

[TITLE]: \texttt{[title]}

[PARENT\_COMMENT]: \texttt{[parent]}
    
\end{quote}

\paragraph{Prompt for Classification (Few-shot).}

\begin{quote}
Determine the type of fallacy in the given [COMMENT]. The fallacy would be one of in the [LOGICAL\_FALLACY] list. Utilize the [TITLE] and [PARENT\_COMMENT] as context to support your decision, and provide an explanation of the reasoning behind your determination.

Here are some examples:

[TITLE]: \texttt{[title 1]} [PARENT COMMENT]: \texttt{[parent comment 1]} [COMMENT]: \texttt{[comment 1]} [OUTPUT]: \texttt{[label 1]} [EXPLANATIONS]: \texttt{[explanation 1]}

[...]

[TITLE]: \texttt{[title 6]} [PARENT COMMENT]: \texttt{[parent comment 6]} [COMMENT]: \texttt{[comment 6]} [OUTPUT]: \texttt{[label 6]} [EXPLANATIONS]: \texttt{[explanation 6]}

[COMMENT]: \texttt{[comment]}

[LOGICAL\_FALLACY]" \texttt{[fallacy]}

[TITLE]: \texttt{[title]}

[PARENT\_COMMENT]: \texttt{[parent]}
    
\end{quote}

\paragraph{Prompt for Classification (COT).}

\begin{quote}
Determine the type of fallacy in the given [COMMENT]. The fallacy would be one of in the [LOGICAL\_FALLACY] list. Utilize the [TITLE] and [PARENT\_COMMENT] as context to support your decision, and provide an explanation of the reasoning behind your determination.

Let's think step by step. First, answer these questions:
\begin{itemize}
    \item What are the differences between fallacies in the [LOGICAL\_FALLACY] list?
    \item For each fallacy type, are there any common patterns in the fallacious sentence?
\end{itemize}

Then, use the answers to these questions to determine the type of logical fallacy in the given [COMMENT].

[COMMENT]: \texttt{[comment]}

[LOGICAL\_FALLACY]" \texttt{[fallacy]}

[TITLE]: \texttt{[title]}

[PARENT\_COMMENT]: \texttt{[parent]}
    
\end{quote}

\section{Data Diversity\label{append:data_diversity}}

\paragraph{\dataset covers diverse topics.}
Table~\ref{tb:topic_statistics} shows the proportions of each topic in \dataset. 
As each news article may have multiple topics, the summation of each column may exceed 100\%. 
The result indicates that most of the news we collected is related to \emph{international relations}, \emph{women rights}, \emph{police brutality}, \emph{COVID/health issue}, \emph{freedom of speech}, \emph{digital rights}, and \emph{East Asian politics}.

\begin{table}[t]
    \centering
    \small
    \begin{tabular}{lccc}
        \toprule
        Topic & Train & Dev & Test \\
        \midrule
        Protest & 2.9\% & 3.1\% & 3.0\% \\
        International Relations & 11.5\% & 12.4\% & 11.9\% \\
        Race Issue & 4.9\% & 4.7\% & 4.5\% \\
        Women Rights & 9.3\% & 10.1\% & 10.4\% \\
        Russo-Ukrainian War & 7.7\% & 9.3\% & 6.0\% \\
        Environmental Issue & 8.8\% & 10.1\% & 7.5\% \\
        Gender Issue & 3.8\% & 3.1\% & 4.5\% \\
        Human Rights & 1.8\% & 1.6\% & 3.0\% \\
        Drug Issue & 0.2\% & 0.0\% & 0.0\% \\
        Police Brutality & 16.8\% & 14.0\% & 14.9\% \\
        Immigration / Refugees & 7.1\% & 5.4\% & 6.0\% \\
        COVID / Health Issue & 12.6\% & 13.2\% & 9.0\% \\
        Legislation & 6.2\% & 7.0\% & 6.0\% \\
        Freedom of Speech & 14.8\% & 11.6\% & 14.9\% \\
        Election & 6.2\% & 4.7\% & 3.0\% \\
        Sustainability & 5.1\% & 4.7\% & 6.0\% \\
        Religious Conflict & 2.0\% & 2.3\% & 1.5\% \\
        Political Debates & 4.0\% & 3.9\% & 4.5\% \\
        U.S. Politics & 0.2\% & 0.0\% & 3.0\% \\
        Digital Rights & 11.5\% & 14.0\% & 11.9\% \\
        East Asian Politics & 9.7\% & 7.8\% & 9.0\% \\
        \bottomrule
    \end{tabular}
    \caption{Proportions of different topics in each split. The distribution of topics remains consistent across all splits, with each topic maintaining a similar proportion regardless of the split.}
    \label{tb:topic_statistics}
\end{table}

\paragraph{\dataset contains comment sections with diverse thread structures.}
To analyze the structure of discussion threads in \dataset, we categorized the structures into four types:
\begin{itemize}[nosep]
    \item \textbf{Flat:} Every comment directly responds to the news article.
    \item \textbf{Single Conversation:} Only one comment received one or more replies.
    \item \textbf{Multiple Conversations:} Several comments received replies, but none of these replies received their own responses (no second-layer responses).
    \item \textbf{Complex:} Any structure that does not fit into the above categories.
\end{itemize}
We calculated the diversity of structures using the evenness index
$J$, proposed by \citet{PIELOU1966131}:
\begin{align}
    J=H/\log S
\end{align}
where 
\begin{align}
    H=-\sum_i p_i \log p_i
\end{align}
$H$ is the Shannon Diversity Index~\citep{Shannon_1948}, $S$ is the total number of unique structures, and $p_i$ is the proportion of a unique structure within its category. 
The value of $J$ ranges from 0 to 1, with higher values indicating greater evenness in structure diversity.
Table~\ref{tb:thread_structure} shows the statistics for each thread structure type in \dataset.
In total, \dataset had 347 unique thread structures, most of which were of Single Conversation.
The diversity of thread structures was high. 


\begin{table}[t]
    \centering
    \small
    \begin{tabular}{l@{\hskip4pt}ccc}
        \toprule
        Type & \makecell[b]{\# Unique\\ Structures} & \# Articles & \makecell[b]{Evenness\\ ($J$)}\\
        \midrule
        Flat & 5 & 26 & 0.51\\
        Single Conversation & 134 & 312 & 0.93\\
        Multi Conversation & 149 & 246 & 0.96\\
        Complex & 59 & 64 & 0.99\\
        \midrule
        Total & 347 & 648 & 0.95\\
        \bottomrule
    \end{tabular}
    \caption{Statistics of the thread structure. The 648 comment threads we collected formed 347 unique structures, with the majority falling under the category of `Multi Conversation'.
    }
    \label{tb:thread_structure}
\end{table}

\begin{figure*}[t]
    \begin{subfigure}{0.45\textwidth}
    \includegraphics[width=\textwidth]{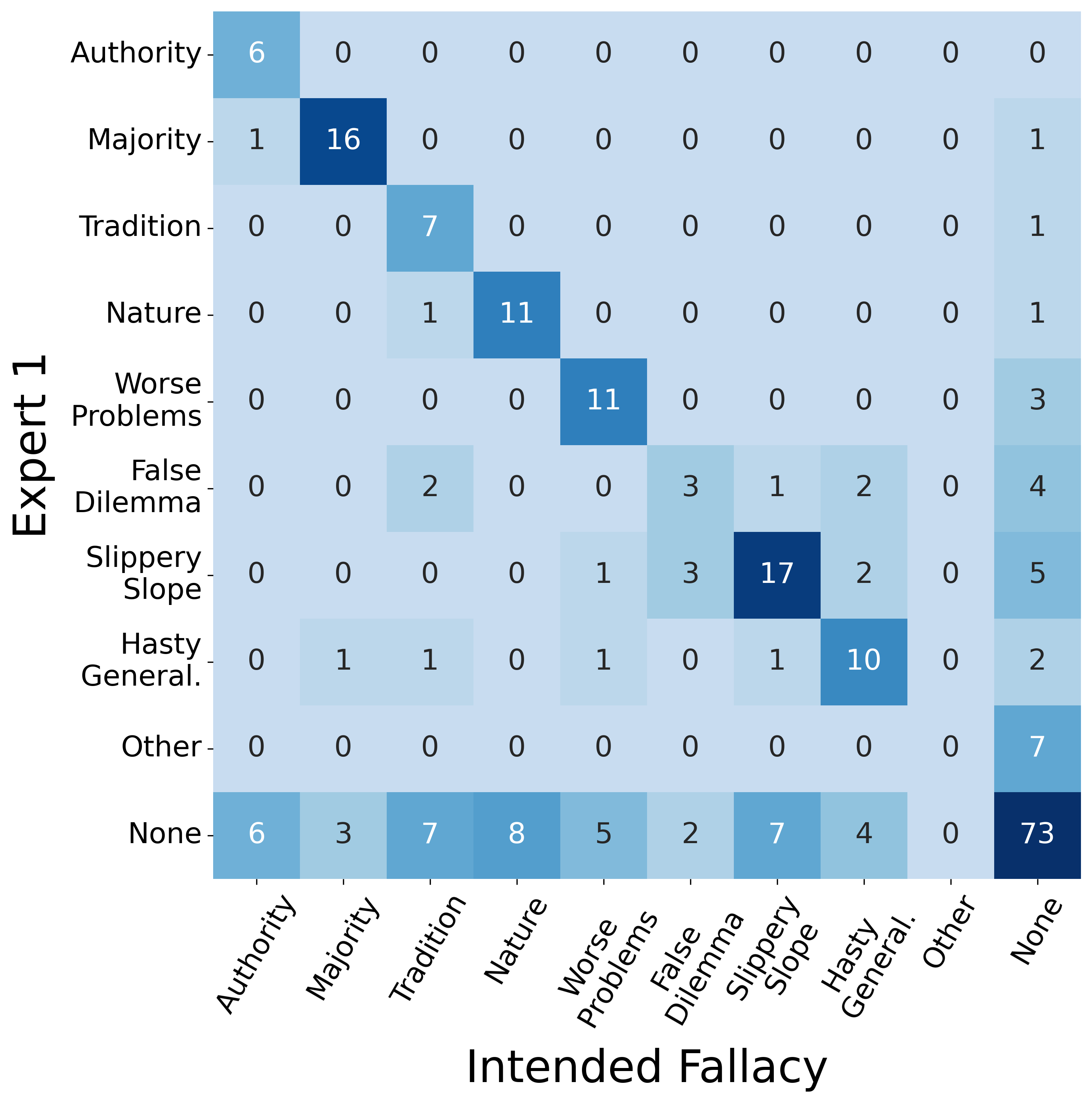}
    \caption{Expert 1 vs. labels (\dataset).}
    \end{subfigure}
    \hfill
    \begin{subfigure}{0.45\textwidth}
    \includegraphics[width=\textwidth]{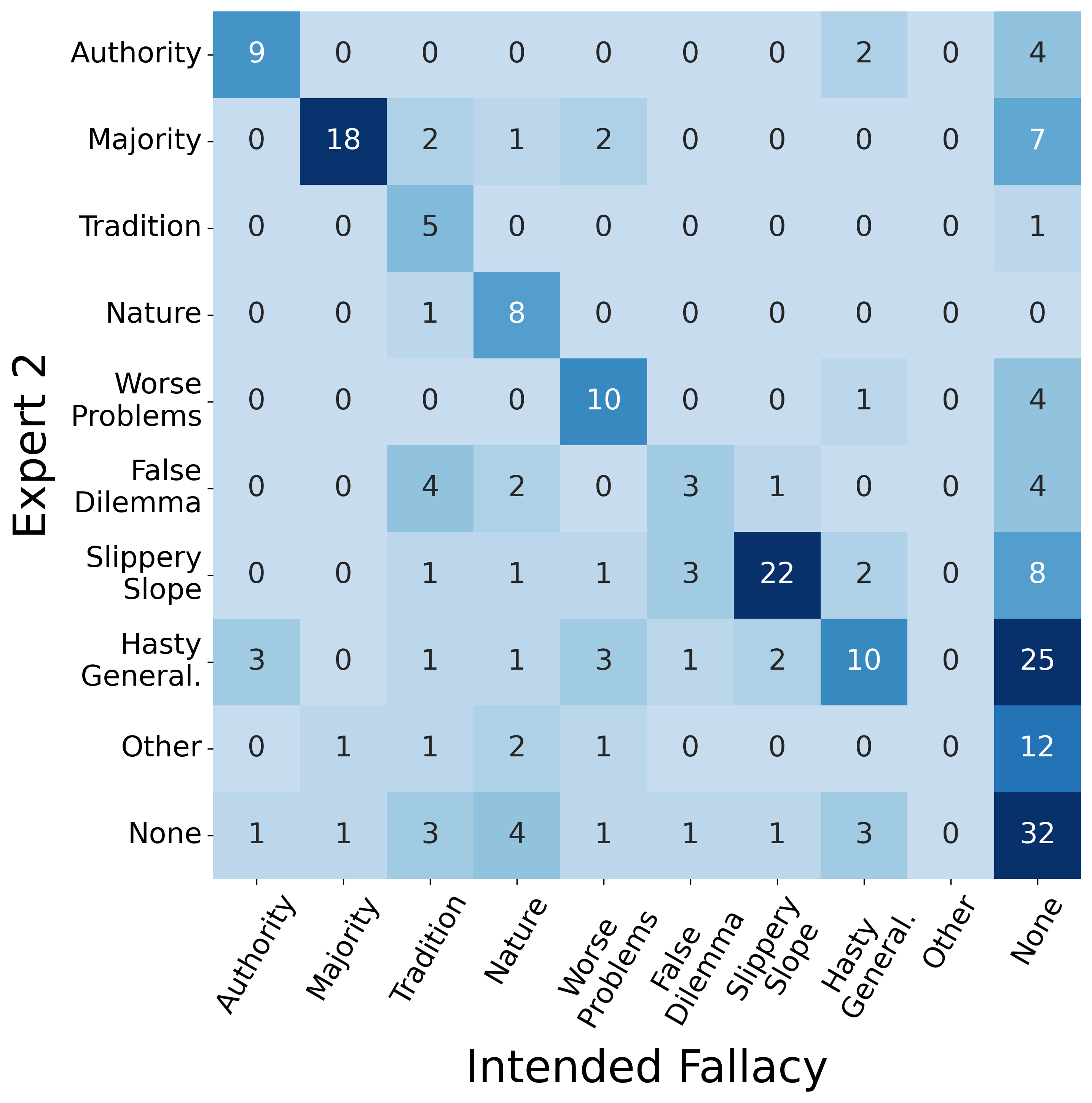}
    \caption{Expert 2 vs. labels (\dataset).}
    \end{subfigure}
    \begin{subfigure}{0.45\textwidth}
    \includegraphics[width=\textwidth]{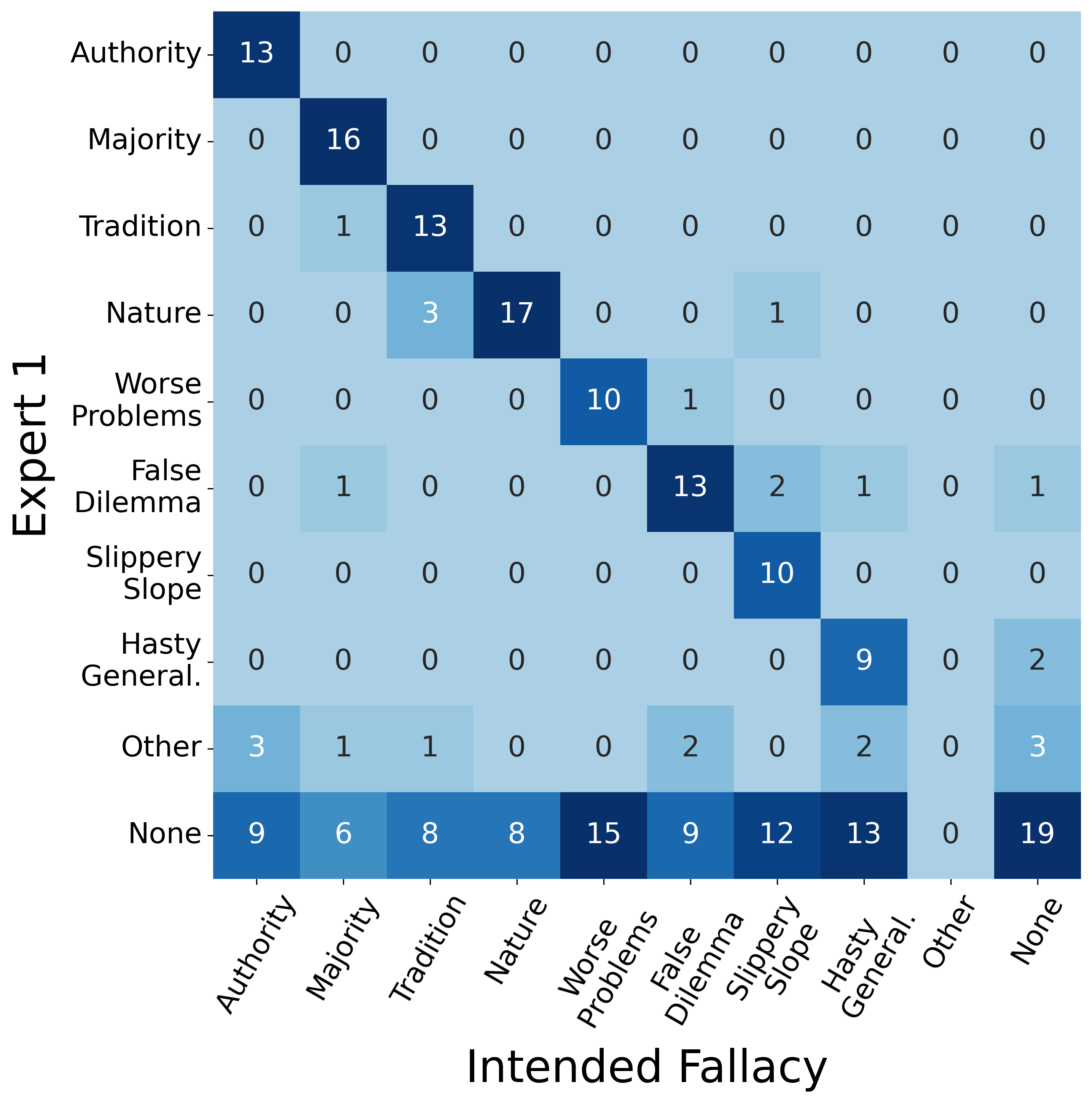}
    \caption{Expert 1 vs. labels (Reddit).}
    \end{subfigure}
    \hfill
    \begin{subfigure}{0.45\textwidth}
    \includegraphics[width=\textwidth]{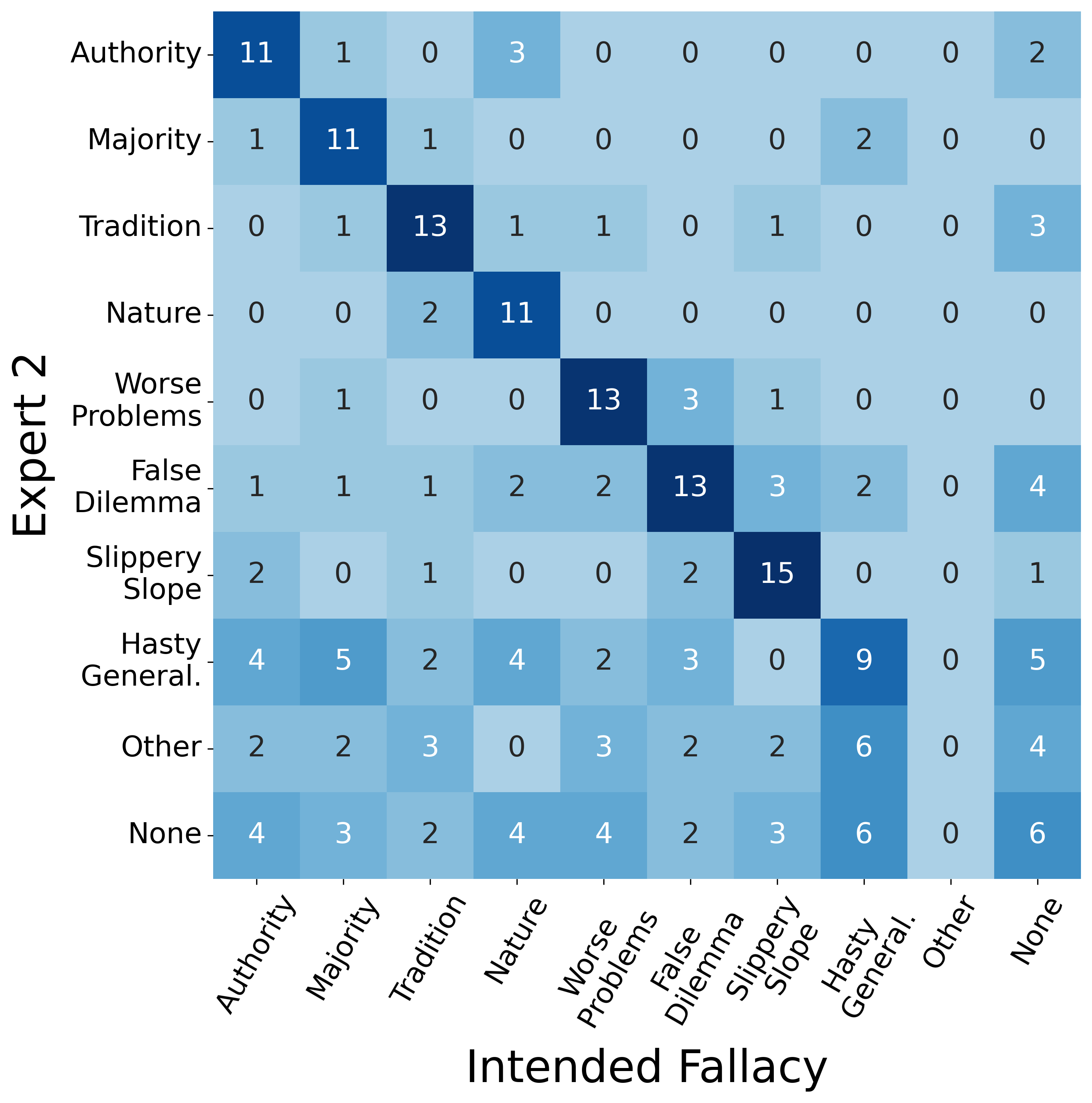}
    \caption{Expert 2 vs. labels (Reddit).}
    \end{subfigure}
    \begin{subfigure}{0.45\textwidth}
    \includegraphics[width=\textwidth]{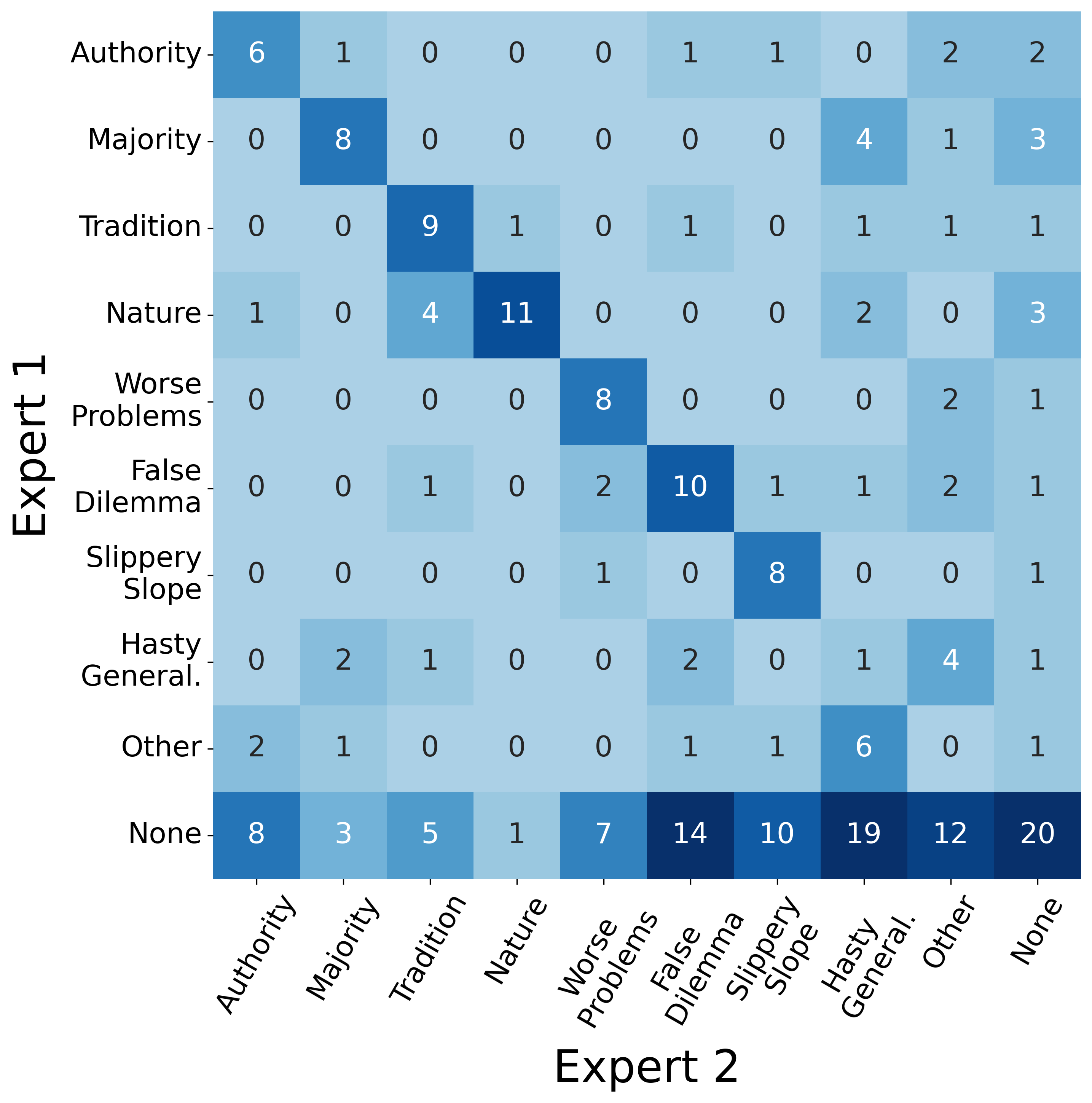}
    \caption{Expert 1 vs. expert 2 (Reddit).}
    \end{subfigure}
    \hfill
    \caption{
    The confusion matrix of the annotation agreement.
    }
    \label{fig:confusion_matrix}
\end{figure*}

\section{Annotation Agreement}\label{append:annotation_agreement}

Figure~\ref{fig:confusion_matrix} shows the confusion matrices between experts annotation and labels for both \dataset and Reddit datasets, as well as the confusion matrix between two experts annotation on the Reddit datasets.

\section{Experimental Details}\label{append:experimental_details}

We had two different versions of BERT and NLI models. One was fine-tuned on the Reddit dataset, the other was fine-tuned on \dataset. 
We fine-tuned them with default hyperparameters set in the original paper, \ie, \citet{Sahai_Balalau_Horincar_2021} and \citet{Jin_Lalwani_Vaidhya_Shen_Ding_Lyu_Sachan_Mihalcea_Schoelkopf_2022}, respectively. Both models were fine-tuned on a server with an A100 GPU. The training took less than 2 hours for each settings. We ran Llama3 on the same server with Ollama,\footnote{Ollama: \url{https://ollama.com/}} a package that allows us to run open-weight LLMs with 4-bits quantization on a local server. The inference took 5 to 20 seconds for each instance, depending on the prompt and the input.

\section{Additional Results on NYT}\label{ap:additional_nyt_res}

To increase the reliability of the NYT annotation, we hired another expert to annotate 250 NYT comments sampled from the annotation set. The overall Cohen’s kappa score between two experts is 0.22, echoing our finding in Sec~\ref{sec:analysis} that it is hard to obtain high IAA in logical fallacy annotation, and that logical fallacy detection in the wild is hard.

Table~\ref{tb:nyt_detection_new_res} shows the performance of different models on the 250 samples. We considered both union and intersection labels, where the former one considered a borderline case as fallacy while the latter one considered it as non-fallacy. The result suggests that models fine-tuned on \dataset generally outperform those trained on Reddit, aligning with the result we showed in Sec~\ref{sec:nyt_res}.

\begin{table}[]
\small
\centering
\begin{tabular}{lcccc}
\toprule
Model & \makecell[c]{Train On / \\ Prompt} & P  & R  & F  \\ \midrule
\multirow{2}{*}{BERT} & Reddit & 84 / 33 & 66 / 62 & 74 / 43 \\
& \dataset & 90 / 37 & 58 / 57 & 70 / 45 \\ \midrule
\multirow{2}{*}{NLI} & Reddit & 81 / 36 & {91} / {95} & 86 / 52 \\
& \dataset & 88 / 40 & 59 / 63 & 70 / 49 \\ \midrule\midrule
\multirow{3}{*}{GPT-4o} & zero-shot & {92} / {50} & 69 / {95} & {79} / {65} \\
& few-shot & {95} / {53} & 46 / 60 & 62 / 56 \\
& COT & 90 / 40 & 82 / {88} & {86} / 55 \\ \midrule
\multirow{3}{*}{Llama3} & zero-shot & 92 / {46} & {53} / 64 & 68 / {54} \\
& few-shot & 83 / 36 & {87} / {89} & 85 / 51 \\
& COT & 86 / 44 & {92} / 72 & 73 / 54 \\ 
\bottomrule
\end{tabular}
\caption{The result of fallacy detection on 250 NYT samples labeled by two annotators, aggregated in two ways: union and intersection.
The left/right numbers are scores with union/intersection labels, where the former one considered a borderline case as fallacy while the latter one considered it as non-fallacy.
}
\label{tb:nyt_detection_new_res}
\end{table}



\end{document}